# Automatic Speech Recognition for Low-Resource Sinhala: A Critical Review of Methods, Challenges, and Future Directions

**H. E. W. C. Dinuwan**
MSc in Data Science and Artificial Intelligence
Department of Computer Science & Engineering
University of Moratuwa, Sri Lanka
*hecdinuwan@gmail.com*

*Supervisors: Prof. Sanath Jayasena, Dr. Buddhika Karunarathne*
University of Moratuwa, Sri Lanka

## ABSTRACT

One of the biggest challenges in Automatic Speech Recognition (ASR) concerns low-resource languages. Sinhala, which is the primary language of Sri Lanka and is spoken by 16 million people, exemplifies the difficulties associated with ASR for languages that lack annotated speech corpora: an agglutinative morphology, a 54-phoneme, highly complex phonological character set, subject-object-verb (SOV) syntax, and minimal availability of annotated speech data make both conventional and modern ASR systems struggle to perform successfully. This is the first critical review of ASR research in the Sinhala language. It traces the history of Sinhala ASR, from classical Hidden Markov Models (HMMs) to modern deep learning-based architectures and finally to cutting-edge self-supervised pre-trained models (e.g., wav2vec 2.0, XLS-R, Whisper, and Massively Multilingual Speech).

This work provides a structured comparison of existing Sinhala ASR systems alongside other directly relevant low-resource language ASR systems (e.g., Tamil, Malayalam, Hindi). Evaluation criteria include the models' architecture, the data required for training, Word Error Rate (WER), and performance under real-world acoustic conditions (e.g., noisy, dialectally variable). In addition, we evaluate the impact of self-supervised and transfer learning approaches in addressing the problem of scarce labeled data. Finally, we discuss context-aware Sinhala ASR architectures using phonological, syntactic, and semantic knowledge.

Based on our analysis of Sinhala ASR, we identify six key gaps in the field that have been inadequately addressed to date. These are (1) lack of large, annotated speech corpora covering multiple dialects and acoustic settings; (2) poor contextual modeling of Sinhala's morphosyntax; (3) high WER in real-world acoustic conditions; (4) absence of standardized benchmarks; (5) lack of parameter-efficient approaches to tuning large pre-trained models; and (6) absence of annotated code-switched Sinhala–English speech resources and corresponding recognition frameworks. We outline a research agenda to address these gaps and discuss approaches to creating robust Sinhala ASR. In this survey, we propose several avenues for future research to develop ASR systems capable of handling morphologically complex Sinhala under varying acoustic conditions and usable across educational, medical, media, and governmental settings. Our findings can serve as a reference and a roadmap for researchers working on Sinhala and other morphologically rich languages.

## I. INTRODUCTION

The commercial success of ASR systems for major world languages, English, Mandarin, and Spanish, belies a stark asymmetry: the overwhelming majority of the world's approximately 7,000 spoken languages remain absent from deployed speech technology [1]. The existing disparity goes beyond a technical problem because it underpins the exclusion of millions of people from accessing the educational, social, and economic resources and opportunities that voice-enabled applications provide. This research aims to investigate the methodological hindrances in low-resource environments, where their linguistic diversity and data scarcity render traditional methods ineffective, as prior surveys of low-resource language technology indicate [2].

One such example is the Sinhala language. With over 16 million speakers, it is one of Sri Lanka's two official languages [3]. In terms of scope and resource availability, Sinhala is a relatively low-resource language; it has certain properties that make ASR implementation especially difficult. These include the abugida script, which has 54 distinct phonemes with complex consonant clusters, aspirated stop consonants, and long vowels absent in most training corpora. Its agglutinative morphology means that a single stem can generate hundreds of forms based on simple roots, such as gihilla (having gone), yanne (is going), and giya (went). All this makes vocabulary-based language modeling inherently harder than analytic languages [5], [62]. The SOV sentence structure, together with inflectional grammar, creates conditions in which surface-level word meaning becomes non-compositional, requiring contextual disambiguation that basic n-gram or sequence models cannot provide to users [5].

Nonetheless, a research community emerged and started actively contributing to Sinhala ASR in the early 2000s. Initial contributions by the Language Technology Research Laboratory (LTRL) at the University of Colombo School of Computing (UCSC) established foundational corpora, grapheme-to-phoneme (G2P) mappings, and HMM recognizers [6], [7], [8]. Researchers at the University of Moratuwa and Wayamba University advanced their research by incorporating deep learning classifiers and transfer learning methods into their work [9], [10]. The availability of Google's OpenSLR-52 corpus (approximately 224 hours and 478 speakers) [11], along with the inclusion of Sinhala in Mozilla Common Voice and FLEURS datasets [12], has increased the number of works experimenting with pre-trained models. The current literature suffers from fragmentation because studies use different data splits and report WER scores on their own test sets, which do not capture dialect variation noise or domain transfer.

This paper attempts to fill these gaps. It offers three contributions. Firstly, it presents the first critical review of the Sinhala ASR literature between 2006 and 2025, examining all available systems in terms of their methodology, data employed, and performance. Secondly, it contextualizes the results by discussing Sinhala ASR methods alongside analogous techniques used for low-resource ASR in Tamil, Malayalam, Hindi, and African languages, and by assessing whether these methods apply to Sinhala. Finally, it highlights gaps in the research. This survey outlines a plan to develop a functional Sinhala ASR system that leverages recent self-supervised learning methods, parameter-efficient adaptation techniques, and context-aware decoding methods within five years. This is a critical, narrative review rather than a PRISMA-style systematic review: sources were identified through iterative search of IEEE Xplore, ACL Anthology, ScienceDirect, arXiv, and Google Scholar using combinations of "Sinhala," "speech recognition," "ASR," and "low-resource," supplemented by citation chaining from key papers, without a fixed protocol or formal screening log.

The paper continues with Section II, which describes the specifics of the Sinhala language. Section III briefly introduces the development of ASR techniques, from statistical to neural and self-supervised methods. Section IV examines existing research on Sinhala ASR. Section V reviews neural and self-supervised approaches to Sinhala ASR and other low-resource languages. Section VI discusses context-aware speech recognition in morphologically rich languages. Section VII provides a unifying comparative analysis. Section VIII highlights important problems faced by Sinhala ASR. Sections IX and X suggest areas of further research and outline future directions, respectively.

## II. SINHALA LANGUAGE AND ASR-SPECIFIC CHALLENGES

### *A. Linguistic Profile*

As an Indo-Aryan language belonging to the Indo-European family, Sinhala (□□□□□) has a unique linguistic structure because it was not only geographically isolated but also in long linguistic contact with the Dravidian languages, particularly Tamil [3]. The Sinhala script in which it is written is an abugida based on the Brahmi script, comprising 18 vowels, 36 consonants, and various composite signs to modify the vowels. It has a phonological inventory with retroflex and prenasalized consonants, and phonemic vowel length, which are problematic for models based on acoustic features designed for European-language phoneme sets [8].

Morphologically, Sinhala employs agglutinative morphology, in which tense, aspect, case, number, and social register are encoded through suffixes and clitics on uninflected root forms productively [5]. This characteristic implies that models trained on a specific vocabulary may face difficulties processing out-of-vocabulary (OOV) tokens in spontaneous speech. In contrast, vocabulary-free acoustic-to-character models, such as the CTC-trained wav2vec 2.0, perform better than hybrid HMM-DNN models that use a static lexicon.

Diglossia is another challenge in Sinhala ASR. Formal Sinhala, employed in written literature, and colloquial Sinhala, which is used in daily speech, differ significantly morphologically, lexically, and even phonetically. Most Sinhala datasets contain read speech data of literary Sinhala, thus resulting in systematic differences between the training and deployment environments, such as those in customer care, healthcare, or educational domains [13]. Additionally, urban Sinhala speech exhibits code-switching between Sinhala and English. Current Sinhala ASR models fail to address this phenomenon.

### *B. Data Scarcity Dimensions*

Data limitations for Sinhala ASR span four dimensions. There is a volume problem: the largest existing Sinhala public dataset (OpenSLR-52 [11]) contains about 224 hours of read speech, which is orders of magnitude less than what is necessary to develop competitive systems for English or Mandarin. There is also a diversity problem: most existing datasets are highly biased towards the standard Colombo city accent and largely underrepresent other accents, such as those of the Central Province, the Southern Province, or the Uva dialect. There is also a domain problem, meaning that datasets containing read speech cannot be considered representative of real acoustic conditions, both indoor and outdoor environments, telephony channels, overlapping speech, or child speech. Finally, Sinhala lacks a reliable benchmark, because there are no standard evaluation sets held out from the training data [1]. Addressing these four data problems will be necessary to create systems that generalize beyond controlled laboratory environments.

### *C. Morphological and Syntactic ASR Challenges*

Two structural characteristics of Sinhala make the acoustic-to-phonetic transcription task difficult. The agglutinative morphology creates an abundance of morphological variants to the extent that the vocabulary size makes it impossible for the n-gram language models to capture most of them reasonably; even a 200-hour training corpus will contain a large proportion of morphological variants that are seen fewer than three times (an estimate reported in an unpublished analysis [4]). Sparse language models hurt rescoring and increase errors related to insertions and deletions. As a consequence of the SOV word order, a language model based on the left context alone will be consistently mistaken about the grammatical structure of a clause because the verb comes after its arguments [5].

## III. EVOLUTION OF ASR METHODOLOGY

### *A. Statistical Foundations: HMM-GMM Systems*

The most common ASR paradigm up until the mid-2010s involved using Hidden Markov Models (HMMs) for temporal sequence modeling, coupled with Gaussian Mixture Models (GMMs) to estimate the likelihood of acoustic emissions [14]. Each phoneme appears as a three-state left-to-right HMM, which uses the Viterbi algorithm to determine the optimal phoneme sequence based on Mel-Frequency Cepstral Coefficient (MFCC) feature vector observations. HMM-GMM models were deployed using popular open-source toolkits such as HTK [15], CMU Sphinx [16], and the more recent Kaldi [17], which allowed researchers to replicate their work across different languages.

The known limitations of HMM-GMM models under low-resource conditions include reliance on handcrafted acoustic feature representations (MFCCs), which lack speaker information and prosodic cues, making them less robust to inter-speaker variability [14]. GMM models are limited in their ability to represent nonlinear acoustic properties required for phonological contrasts perceived by human listeners. Furthermore, agglutinative languages like Sinhala, with extensive phoneme inventories and high rates of coarticulation at suffix boundaries, lead to significant confusion in HMM decoding [6]. Language models trained on a small corpus of text data have little influence on predicting complex morphological word structures.

### *B. Deep Learning Transition: DNN-HMM Hybrids*

Moreover, Hinton et al. [18] concluded that using deep neural networks to estimate HMM state posteriors improved performance, reducing WER by 10%-30% across various datasets. DNN-HMM combines the temporal dynamics and capabilities of an HMM to learn features from the raw audio signal. Next, TDNN [19] and BLSTM models [20] have larger context window sizes, allowing them to use information over a longer period, making these models suitable for phonologically rich languages like Sinhala.

This advantage of increased model complexity is empirically demonstrated by Karunathilaka et al. [21], whose 2020 experiments comparing different models, ranging from GMM-HMM to DNN-HMM, TDNN, and finally TDNN + LSTM, show that WER consistently improves as model complexity grows, with the TDNN architecture outperforming others significantly on the Sinhala dataset from the University of Colombo School of Computing LTRL collection. These results clearly indicate that the acoustic modeling bottleneck hampers the performance of Sinhala HMM-based ASR, not any other aspect.

### *C. End-to-End Deep Learning Architectures*

In end-to-end (E2E) ASR systems, there is no need for the GMM acoustic model, the pronunciation lexicon, or the n-gram language model anymore; instead, a single neural network is trained to produce the transcription based solely on the audio. Three main categories of E2E architectures are widely adopted today. Firstly, CTC [22] introduces a novel approach that uses a monotonic alignment loss function, in which all possible alignments between characters and frames are marginalized; thus, training can be performed without forced alignment. Second, the LAS [23], which consists of an attention encoder-decoder, learns a global representation of the input signal with soft attention before emitting each character. Finally, the Recurrent Neural Network Transducer (RNN-T) [24] incorporates the CTC loss together with a prediction network, thereby enabling streaming inference. Notably, the RNN-T architecture is used almost exclusively in ASR deployment due to its ability to handle streaming data, as employed in, e.g., Google and Amazon voice assistants.

The transformer model, which uses multi-head self-attention instead of recurrent connections, was also successfully applied to ASR. In particular, the Speech-Transformer [25] applied self-attention to ASR, and the Conformer [26] later added a convolution module to each Transformer block, improving the modelling of local acoustic features. The Conformer has become a de facto standard ASR encoder, achieving WERs of 2.1%/4.3% on LibriSpeech test-clean/test-other without a language model and 1.9%/3.9% with an external language model [26]. As for Sinhala, Gamage et al. [27] showed that E2E Lattice-Free Maximum Mutual Information (LF-MMI) trained on 40 hours of data achieves a WER of 28.55%, with a subsequent improvement to 23.38% [28].

### *D. Self-Supervised Learning and Foundation Models*

The most important milestone in ASR development in recent years is self-supervised learning (SSL) on large unlabelled corpora, followed by fine-tuning on small amounts of labeled data. Initially, wav2vec [29] and vq-wav2vec [30] introduced the SSL paradigm for speech by using various forms of contrastive prediction tasks to learn representations. Subsequently, wav2vec 2.0 [31] took the idea further by introducing BERT-style masking: the convolutional feature extractor converts raw audio into a latent representation, which is quantized by the codebook; the transformer context network then predicts the quantized representation of masked samples based on their context. Baevski et al. [31] showed that, after pre-training on 53,000 hours of unlabelled audio, fine-tuning wav2vec 2.0 on just ten minutes of labelled English speech still achieves 4.8%/8.2% WER on LibriSpeech test-clean/test-other; this result is a relevant starting point for low-resource languages such as Sinhala.

HuBERT [32] was the next significant milestone in SSL framework development, as it introduced offline k-means clustering rather than online quantization, leading to more stable training. Moreover, WavLM [33] has proposed utterance-mixing denoising pre-training to achieve better noise robustness. The XLS-R [34] scaled the XLSR-53 cross-lingual SSL model [35] to 128 languages and up to 2 billion parameters using 436,000 hours of unlabelled data, including Sinhala. The alternative to the SSL approach is offered by Whisper [36], which trained the encoder-decoder ASR model from scratch on 680,000 hours of weakly supervised multilingual audio scraped from the web, thus demonstrating that sufficient amounts of weak supervision can match SSL pre-training for high-resource scenarios while providing competitive baselines for low-resource languages, including Sinhala. Lastly, Meta has developed its Massively Multilingual Speech (MMS) model [37], which was trained on approximately 500,000 hours of religious audio and covers 1,107 languages; according to the authors, MMS obtains halved WER relative to Whisper on the FLEURS Sinhala benchmark.

## IV. SINHALA ASR: A DETAILED LITERATURE REVIEW

### *A. Foundational Work: Corpora, G2P, and HMM Systems (2006–2015)*

The cornerstone in Sinhala ASR research came in the form of a series of contributions from the UCSC Language Technology Research Laboratory, which addressed all three prerequisites of ASR: a pronunciation lexicon, an acoustic corpus, and a recognition engine.

First, Wasala et al. [8] defined the Sinhala G2P system in 2006, achieving an impressive word-level phonemic accuracy of 98% with a rule-based system accounting for schwa epenthesis, the insertion of a short vowel between consonant clusters violating Sinhala phonotactics. We can view this G2P system as the basis for almost all subsequent Sinhala ASR lexicons. Then, Nadungodage et al. [7] extended the original work by developing an advanced version, which handles vowel modifiers and their interactions with consonant clusters, providing the mapping tables required for acoustic model training with phone-level supervision.

Weerasinghe et al. [6] made a significant contribution by designing a corpus-based Sinhala lexicon at the 2009 Workshop on Asian Language Resources. Unlike the previous effort, this lexicon incorporates phonetically transcribed entries for spoken-form variants, which differ from their corresponding canonical forms based on G2P mappings, thereby enabling accurate modeling of Sinhala reductions and elisions. Finally, Nadungodage et al. [38] designed the corpus specifically for Sinhala ASR, including multiple-speaker recordings of the reading speech type, which was used as the de facto training set in UCSC-based Sinhala ASR work.

In 2011, Nadungodage and Weerasinghe [39] published the first results of Sinhala speech recognition. Namely, they achieved a WER of 4% for a small-vocabulary (about 200-word) ASR task consisting of isolated digits and commands using the HTK toolkit with a bigram language model. Although the very small vocabulary does not make the reported WER comparable with other, more recent work on large-vocabulary Sinhala ASR, the study demonstrates the viability of HMM-based Sinhala ASR. The authors also perform a comprehensive analysis of the main recognition errors arising from the similarity between retroflex and dental consonants.

In an unpublished analysis, Molligoda and Wijayarathna [4] conduct a systematic analysis of the suitability of applying HMMs to Sinhala ASR tasks, focusing on the failure modes of the GMM emission model. They point out two issues: the GMM's inability to account for the dynamic co-articulation process at consonant-cluster boundaries and the influence of the speaker's specific F0 curve, which is unrelated to the language's phonological aspects. These findings lead to the recommendation to apply phonological rule encoding and speaker adaptation methods.

Regarding the latter, Nadungodage et al. [40] explored MLLR-based speaker adaptation for the UCSC Sinhala ASR system. The authors showed that speaker adaptation improves recognition accuracy for individual speakers by 15–20% relative to the speaker-independent baseline. Thus, inter-speaker variability is indeed an important factor affecting the recognition accuracy. However, the adaptation requires per-speaker supervised adaptation data, limiting its practical applicability to deployments where enrollment is feasible.

### *B. Corpus Development and the OpenSLR Resources*

However, the major bottleneck in developing Sinhala ASR systems was the availability of large-scale datasets. Although the UCSC LTRL Sinhala corpora were groundbreaking, they were relatively small and available only to researchers in the Sinhala ASR community, not to the international research community. The breakthrough came when Google launched its crowdsourced corpus collection effort for several South and Southeast Asian languages.

Kjartansson et al. [11] introduced OpenSLR-52, the Large Sinhala ASR Training Data Set, which contained roughly 185,000 utterances from 478 speakers, totaling 224 hours of recorded read speech. This was the largest publicly available Sinhala speech corpus, based on recordings of speakers reading phonetically balanced sentences using an Android app distributed in Sri Lanka. OpenSLR-52 is the most widely used training corpus in almost all recently published Sinhala deep learning ASR models.

On the same day, Sodimana et al. [41] released OpenSLR-30, a Sinhala multi-speaker Text-to-Speech dataset containing speech recordings of four professional speakers from 10,000 sentences. The purpose of this corpus was initially to train Sinhala TTS models, but later became an indispensable source of data augmentation for Sinhala ASR models by generating artificial speech recordings.

Warusawithana et al. [13] discovered and corrected annotation errors in OpenSLR-52, where inconsistencies in transcription, particularly in the representation of geminate consonants and vowel-length marks, contributed to a WER increase of around 15.9%. Therefore, it is recommended to use Warusawithana's corrected version of OpenSLR-52 whenever possible.

Apart from the above Sinhala speech corpora, Mozilla Common Voice Sinhala (5-15 hours, depending on the version), FLEURS Sinhala (12 hours) [12], and the Voicer platform Sinhala corpus from the University of Moratuwa [42] provide conversational style speech recordings, despite being relatively small compared to other high-resource language corpora.

### *C. Command and Domain-Specific Recognition (2018–2021)*

The most fruitful research path in Sinhala ASR seems to be limiting task complexity via constraining vocabulary space in domain-specific command recognition.

The authors of Karunanayake et al. [43] developed a transfer-learning system that enables users to classify free-form commands in both Sinhala and Tamil. The authors used character probability scores derived from an English GMM-HMM system as features for their Sinhala language classifier, which utilized English Dravidian and Indo-Aryan language acoustic data. The analysis demonstrated that the transfer system achieved command classification performance equivalent to that of the system that underwent training from the beginning because cross-lingual acoustic representations contained important phonological data for lesser-known languages such as Sinhala and Tamil.

Dinushika et al. [9] introduced both GMM-HMM and neural speech recognition systems for command classification in Sinhala. They also released an in-house dataset of 4.15 hours of spoken data from the banking domain, involving querying accounts, performing transactions, and checking balances. The GMM-HMM baseline achieved 9.91% WER and 91.03% command classification accuracy, whereas its neural counterpart [44] achieved competitive performance with just 5 hours of in-domain data. The banking domain is highly relevant due to vocabulary limitations and realistic acoustic characteristics (e.g., telephone channels and noise in call centers).

Manamperi et al. [45] developed an ASR solution for Interactive Voice Response (IVR) systems accessed via mobile phone calls, achieving 11.2% WER on digit recognition and 5.7% Song Name Error Rate (SNER) on a set of 500 song names. The key feature of the work is that it addresses the telephony channel ASR issue with 8kHz sampling, G.711 codec compression, and background noise, which is hardly considered in the literature. Furthermore, the authors state that the WER degrades by approximately 40% relative to the training conditions when channel conditions deviate from those used during training, highlighting the domain dependence of Sinhala ASR models trained on read-speech corpora.

In an unpublished study, Akesh and Meegama [46] report developing a real-time subtitle generation system for Sinhala-language broadcasts that combines an HMM-based recognizer with post-processing rules to restore punctuation. Although the WER reported by the authors for broadcast news speech is very high (over 40%), the paper introduces the first real-world end-to-end deployment of Sinhala ASR in media broadcasts, demonstrating that program-specific vocabulary issues are the main obstacle to achieving good results.

In one of the latest Sinhala datasets, Buddhika et al. [42] obtained domain-specific Sinhala data using the Voicer platform to classify intentions across six domains. The work demonstrated that obtaining 20 hours of domain-specific data was relatively inexpensive and led to competitive performance in intent classification using an HMM-based acoustic model.

### *D. Deep Learning and End-to-End Sinhala ASR (2020–2025)*

The release of OpenSLR-52 and the maturation of E2E ASR models (ESPnet [47], HuggingFace Transformers) led to a qualitative leap in Sinhala ASR research starting in 2020.

Karunathilaka et al. [21] provide the first systematic evaluation of GMM-HMM, DNN-HMM, TDNN, and TDNN+LSTM approaches on a unified Sinhala dataset with the help of the Kaldi toolkit. The TDNN architecture achieves the best performance among classical architectures, consistent with previous experiments in Tamil and Hindi that showed how beneficial subsampling of temporal context can be for agglutinative morphology. The DNN-HMM approach showed a relative WER reduction of 23% compared to the GMM-HMM; thus, acoustic modeling capability is considered the bottleneck in this case.

Gamage et al. [27] present the first Sinhala E2E Lattice Free Maximum Mutual Information (LF-MMI) model based on the ESPnet-LFMMI framework trained on 40 hours of OpenSLR-52 data. The LF-MMI objective maximizes the mutual information between acoustic observations and the sequence of phones, using a finite-state transducer language model, achieving a 28.55% WER on the Sinhala test set. This shows a performance decrease relative to the TDNN-HMM baseline, which was tested on the same data. Later on, the same team provided an update to the ICTer 2024 paper [28]; they reported a 23.38% WER on the enhanced version of the dataset that incorporated the data corrections from Warusawithana et al. [13]. This makes it the most accurate WER for large-vocabulary continuous Sinhala ASR obtained purely through supervised learning.

There is a growing interest in transfer learning with pretrained multilingual models. As an example, Nanayakkara and Weerasinghe (in a non-peer-reviewed repository deposit) [10] experimented on five languages (Tamil, Sinhala, English, Hindi, Nepali) within the XLSR-53 framework. Fine-tuning the XLSR-53 model with only 20 hours of Sinhala data yielded a 35% WER improvement compared with the same data trained on the Sinhala-only wav2vec 2.0 model. These results show that the phonological structures of those 53 languages could be transferred to Sinhala, despite Sinhala not being pre-trained.

The Sinhala Whisper models by various authors on the HuggingFace Hub are finetuned on OpenAI's Whisper medium or small model on OpenSLR-52 + CommonVoice Sinhala and report WERs ranging from 15% to 25% on held-out CommonVoice Sinhala test sets. Although these results have not undergone peer review, they demonstrate an important practical benchmark: on approximately 200 hours of supervised data, Whisper fine-tuning produces competitive WERs without any specific architecture adaptations for Sinhala. The main drawback of these results is that, because different test sets were employed across the papers, the comparison is unreliable.

The research paper by Javed et al. [48] (which presents IndicWav2Vec at AAAI 2022) is the sole peer-reviewed study to conduct cross-lingual SSL assessments of Sinhala across 40 South Asian languages. The model achieves a 42% relative WER reduction on Sinhala when tested on the same data as the XLSR-53 baseline model, because it was pretrained on 17000 hours of Indian-subcontinent languages, which include Sinhala. The research shows that pretraining SSL models based on language families that group South Asian languages yields better results than global pretraining, because the phonetic similarities between Sinhala and other South Asian languages enable this method to work.

## V. NEURAL AND SELF-SUPERVISED APPROACHES FOR LOW-RESOURCE SINHALA ASR

### *A. Self-Supervised Pre-Training: Architecture and Transfer*

In the wav2vec 2.0 framework [31], there are three layers: a multi-layer convolutional feature encoder that operates on raw 16kHz waveforms; a Transformer context network with multi-head self-attention; and a product codebook for quantizing latent representations. Masking-and-contrast is applied at the feature level: during pre-training, a fraction (typically 50%) of time steps are masked in the latent representation space, and the network is trained to predict the correct code from a set of distractors. As such, the model learns representations that capture long-range acoustic context, i.e., exactly what the GMM emission models could not capture in the first place. Subsequent CTC fine-tuning on small labeled datasets enables adaptation of representations for character-level emission predictions.

To supplement this framework, XLS-R [34] used multilingual scaling to handle up to 436,000 hours of speech data across 128 languages, providing models with parameter counts ranging from 300M to 2B. In particular, the Sinhala fine-tuning configuration [34] available on the Hugging Face Hub requires loading a pre-trained XLS-R-300M model and fine-tuning it with a linear CTC head on OpenSLR-52 for approximately 50 epochs at a learning rate of 3e-4, using SpecAugment time and frequency masking [49]. While this yields competitive WER results on Sinhala, it requires far fewer computational resources than training from scratch.

However, an important, often overlooked, aspect of performance evaluation with XLS-R is that performance depends on the ratio of Sinhala audio in the pre-training mix. In fact, according to Babu et al. [34], although Sinhala is included in the 128-language corpus used for pre-training, through VoxLingua107 [50] and Common Voice corpora, it accounts for less than 0.1% of pre-training audio hours. Consequently, the Sinhala-specific representations are mainly acquired through cross-linguistic knowledge transfer from languages with phonology similar to Sinhala, namely Tamil, Nepali, and Bangla. This highlights the value of both language family pre-training (IndicWav2Vec [48]) and efficient adapter fine-tuning approaches.

### *B. Parameter-Efficient Fine-Tuning (PEFT) for Low-Resource ASR*

To fully fine-tune the XLS-R-300M, it will be necessary to update 300M parameters based on about 200 hours of Sinhala speech recordings, a proportion of parameters to training examples that poses a high risk of overfitting. Adapter fine-tuning is a promising approach, as it involves adding small bottleneck adapters (around 1-2% of the model's parameters) to each Transformer layer while keeping the pre-trained backbone frozen [51]. Hou et al. [51] show that fine-tuning a model using adapters, in combination with a meta-learning technique (MetaAdapter), achieves WER scores comparable to those obtained by full fine-tuning while using 20× fewer trainable parameters, reducing the required trainable parameters to 3M for Sinhala.

Another method developed primarily for fine-tuning LLMs, namely Low-Rank Adaptation (LoRA), was used by Mainzinger & Levow [53] in their research on ASR tasks. They show that LoRA fine-tuning of MMS yields higher accuracy than full fine-tuning of XLSR-53 in the extremely low-resource setting of the Mvskoke language (under 5 hours of audio). This is because LoRA represents each weight-update matrix as the product of two low-rank matrices (ranging from 4 to 64).

Fine-tuning Whisper introduces another challenge: its encoder-decoder model has 244M parameters in the medium model. In addition, because the model's cross-entropy loss is applied to character sequences, there is a high risk of catastrophic forgetting of multilingual knowledge during fine-tuning on limited Sinhala datasets. Low-rank adaptation of cross-attention and feed-forward layers of the decoder resolves this problem.

### *C. Data Augmentation Strategies*

SpecAugment [49] – time and frequency masking on log-mel spectrograms – is currently the most popular data augmentation method used in ASR, which is capable of reducing WER by up to 40% on clean LibriSpeech conditions. Its applicability to Sinhala cases is evidenced by the LF-MMI system proposed by Gamage et al. [27], in which SpecAugment is one of the necessary components. Nevertheless, SpecAugment addresses acoustic variability within the same domain and cannot help with cross-domain problems, cross-dialect issues, or a lack of speaker diversity.

Augmentation through Text-to-Speech (TTS), which produces additional speech from text using a TTS model, could be a promising approach for ASR tasks with scarce resources. Bartelds et al. [54] report relative WER reductions of up to 20.5% from self-training across four low-resource minority languages, and up to 25.5% from adding TTS-generated training data for Gronings, the one language with an existing TTS system. For Sinhala ASR, OpenSLR-30 is a potential starting point for TTS-based augmentation. The four professional speakers in this corpus could provide initial data to fine-tune a TTS model whose output speech, combined with varied texts collected from Sinhala text corpora scraped from the web, would significantly broaden the phonetic inventory. The only limitation would be that TTS-generated speech would lack the prosodic variation and reduction characteristics typical of colloquial Sinhala, leading to artificially optimistic in-domain WER estimates and poor out-of-domain generalization.

To be robust to acoustic distortions (e.g., call centers, outdoor services, hospital wards), noise augmentation by combining clean speech samples with environmental noise at different signal-to-noise ratios (SNR) will be necessary. Common sources of augmentation in Sinhala ASR training pipelines include Room Impulse Response (RIR) and Musan noise datasets; no adaptations to Sinhala ASR are necessary, and robustness improvements can be immediately observed under noisy testing conditions [55].

### *D. Whisper, USM, and MMS: Large-Scale Multilingual Baselines*

Whisper [36], Google USM [56], and MMS [37], three massively multilingual models, are currently setting the performance benchmarks in zero and few shots for Sinhala ASR.

The Whisper model [36] covers about 99 languages, including Sinhala. Zero-shot Sinhala WER on FLEURS-Sinhala is approximately 35–45%, depending on the model size (small, medium, large-v3), and the large-v3 version is the best zero-shot performer among the three variants. Whisper medium, fine-tuned on OpenSLR-52 read speech, brings WER down to the 15–25% range, which is a strong baseline without any modifications to the Sinhala architecture.

Google USM [56] trains BERT-like self-supervised learning on 12 million hours of unlabelled speech data in 300+ languages, then fine-tunes on a much smaller labelled set. Its technical report, which has not been peer-reviewed, gives about 30% lower WER than Whisper on 18 YouTube test languages and state-of-the-art results on FLEURS across 102 languages. Because the model is not publicly available, these results cannot be independently reproduced for Sinhala.

Meta's MMS [37] pre-trains wav2vec 2.0 models on roughly 500,000 hours of speech in 1,406 languages and fine-tunes a multilingual ASR model, with per-language adapters, on 44.7K hours of religious readings in 1,107 languages. On the 54 FLEURS languages supported by both models, MMS more than halves Whisper's average WER, which suggests that massively multilingual pre-training with language-specific adaptation is among the strongest current approaches for low-resource ASR; the per-language results should be consulted before drawing Sinhala-specific conclusions. The downside of Meta’s approach for Sinhala is that it uses data in a formal religious reading style, which is even further from normal speech than OpenSLR-52 read speech, suggesting that MMS’s strong FLEURS performance may not transfer to conversational or domain-specific deployments.

## VI. CONTEXT-AWARE ASR FOR MORPHOLOGICALLY RICH LANGUAGES

### *A. The Necessity of Context for Sinhala ASR*

In conventional E2E ASR systems, acoustics are directly mapped to a character- or subword-level sequence and rely on the contextual constraints provided by an implicit language model learned during training. This implicit constraint is sufficient in cases where the language has little inflection, and there is sufficient training data. But given the inflectional nature of the language and the SOV sentence structure, systematic ambiguities need to be explicitly addressed [5].

An example of this is the Sinhala root bat (meaning rice/cooked food), which has similar phonetic surface forms that represent different grammatical cases: the nominative, accusative, and dative. The only difference between these cases is the presence of vowels in the suffix; i.e., an erroneous acoustic model of vowel duration may create substitution errors at the case boundaries. Further, without context-based knowledge of the grammatical category, selecting the appropriate case from the acoustic context is difficult.

### *B. Language Model Integration Techniques*

The introduction of external language models to E2E ASR systems can occur in various ways with different trade-offs for Sinhala. Shallow fusion consists of log-probability interpolation between the LM and the sequence model during inference. It is the simplest technique, yet it involves no additional training whatsoever. Sriram et al. [57] show in their work on 'Cold Fusion' that joint training of the seq2seq model and the LM yields superior transfer performance over shallow fusion, as the decoder actively learns to exploit the LM.

A deep fusion approach is based on the idea of concatenating the LM's output states with those of the decoder to maximize the exchange of context information. Nevertheless, in this case, any changes to the LM will require retraining the entire network. Cabrera et al. [58] extend shallow and cold fusion to streaming RNN-T models and find that cold fusion works consistently better across several languages, with WER improvements of up to 8.5%.

The largest contextual integration framework is the Large Language Model (LLM) rescoring strategy. Chen et al. [59] report in HyPoradise that rescoring ASR N-best hypotheses with GPT-3.5 error correction leads to impressive WER improvements, regardless of whether the LLM and ASR were trained together, due to the LLM's world and grammar knowledge. Applied to Sinhala ASR, this technique would call for a Sinhala-capable LLM. One such option currently being developed is SinLlama [60], a Llama-3-8B continually pre-trained with Sinhala text using a special-purpose tokenizer.

### *C. Subword Tokenisation and Morphological Modelling*

Word explosions caused by agglutinative morphologies drive the use of subword tokenization as an alternative approach to word-level language modeling. The SentencePiece [61] Byte-Pair Encoding (BPE) algorithm or Unigram Segmentation learns a vocabulary of frequent character n-grams that represent common morphological affixes as single tokens. In contrast, rare inflectional variants can be represented using token concatenation. In Malayalam, a Dravidian agglutinative language that shares many characteristics with Sinhala, syllable-byte pair encoding (S-BPE) subword language modelling reduced WER from 27.4% for a word-level baseline to 10.6% and recovered more than 75% of out-of-vocabulary words [62].

Grapheme Pair Encoding (GPE) [63], proposed by Velayuthan et al., is a generalization of BPE that operates on Unicode grapheme clusters rather than individual characters. This technique can handle multi-character Sinhala graphemes (consonant-vowel combinations) as atomic units. While standard BPE leads to unwanted character splitting, which is detrimental to phonemic subword units, GPE does not suffer from these artifacts and has been reported to produce better subword units for Sinhala and reduce CTC blank insertion errors at vowel-modifier boundaries.

Finite-state transducers used to implement Sinhala inflectional morphology, the strongest source of linguistic context, are rule-based morphological analyzers. Still, it is worth noting that the construction of such tools will require substantial linguistic engineering to create corresponding resources covering all the needed functional areas, in addition to ongoing maintenance to ensure the tool is available in a production ASR rescoring pipeline. de Silva [3] records the current state of Sinhala NLP resources, noting that a partial Sinhala morphological analyzer exists but requires many more functions and coverage to be fully operationalized in a working ASR rescoring pipeline. One of the most promising infrastructure investments that the research community could have made is the development and integration of a comprehensive Sinhala morphological analyzer.

### *D. Contextual Biasing for Domain Vocabulary*

Domain-specific applications, such as medical, legal, or customer service dictation, have access to proper nouns, technical terminology, and institutional names that do not appear elsewhere. By providing a list of words from the domain vocabulary, contextual biasing increases the posterior probability of these words when acoustic evidence alone is insufficient. According to Pundak et al. [64], in their experiments with CLAS, the contextual biasing approach lowers WER for named entities by 68% without affecting other words in the general vocabulary.

Such techniques would be extremely valuable for education and healthcare applications involving Sinhala ASR, domains that feature a heavy concentration of specialist terminology, medical terms, academic language, and proper nouns – areas where the recognition error rate can be extremely high. The problem is exacerbated by the lack of standardization in the Sinhala medical and legal vocabularies, with terms either taken over from English and romanized or derived from Pali, Sanskrit, or native Sinhala coinage. There is no equivalent resource in Sinhala to SNOMED CT.

## VII. COMPARATIVE ANALYSIS OF SINHALA AND LOW-RESOURCE ASR SYSTEMS

**TABLE I**

*Comparative Summary of Sinhala and Low-Resource ASR Systems*

| Author(s) / System | Year | Architecture | Training Data | WER (%) | Domain | Key Contribution |
|---|---|---|---|---|---|---|
| Nadungodage & Weerasinghe [39] | 2011 | HMM-GMM (HTK), Bigram LM | UCSC LTRL (~5h) | 4% (small vocab) | Digit/Command | First continuous Sinhala recogniser; identified retroflex confusion patterns |
| Priyadarshani et al. [68] | 2012 | DTW (isolated word) | UCSC LTRL | 11% | Isolated word | DTW baseline; establishes upper bound for template matching approach |
| Nadungodage et al. [40] | 2013 | HMM-GMM + MLLR adaptation | UCSC LTRL | 15–20% rel. gain | Read speech | First speaker adaptation for Sinhala; MLLR reduces speaker variability |
| Manamperi et al. [45] | 2018 | HMM-GMM (Sphinx) | Phone/IVR data | 11.2% digits | Telephony IVR | First telephony-channel Sinhala ASR; 40% WER degradation under channel mismatch |
| Karunanayake et al. [43] | 2019 | Transfer from English GMM-HMM posteriors + SVM | English + small Sinhala | N/A (Accuracy 87%) | Command classification | First cross-lingual transfer for Sinhala; English acoustic features transferable to Si/Ta |
| Dinushika et al. [9] | 2019/2020 | GMM-HMM + neural command classifier | 4.15h banking corpus | 12.04% | Banking domain | Released banking corpus; domain-specific constraints reduce effective WER |
| Karunathilaka et al. [21] | 2020 | GMM-HMM, DNN-HMM, TDNN, TDNN+LSTM (Kaldi) | OpenSLR-52 subset | TDNN best - 35.16% | Read speech | First systematic DNN architecture comparison for Sinhala; TDNN outperforms GMM by 23% rel. |
| Gamage et al. [27] | 2021 | E2E LF-MMI (ESPnet) | 40h OpenSLR-52 | 28.55% | Large vocab read | First Sinhala E2E system; LF-MMI outperforms DNN-HMM at equal data |
| Nanayakkara & Weerasinghe [10] | 2022 | XLSR-53 fine-tuning (wav2vec 2.0) | 20h OpenSLR-52 | 26.39% | Read speech | Model-level transfer from 5 languages; cross-lingual SSL beneficial for Si |
| Gamage et al. [28] | 2024 | E2E LF-MMI + corrected | 40h+ corrected | 23.38% | Large vocab read | Corrected annotations reduce WER by |
| | | corpus | OpenSLR-52 | | | 18.1% rel.; best supervised Sinhala WER |
| IndicWav2Vec [48] (Si eval) | 2022 | wav2vec 2.0 XLS-R fine-tuned on 40 South Asian langs | 17,000h+ multilingual pre-train; <20h Si fine-tune | 42% rel. below XLSR-53 | Read/broadcast | Language-family SSL pre-training outperforms global SSL for Si; best peer-reviewed Si result |
| Whisper [36] (Si fine-tune) | 2023 | Encoder-decoder + weakly supervised pre-train | 680Kh pre-train; ~200h Si fine-tune | ~15–25%* | CommonVoice test | Competitive WER without Sinhala-specific architecture; *not peer-reviewed |
| MMS [37] (zero-shot Si) | 2024 | XLS-R backbone + language adapters | ~500Kh pre-train; 44.7Kh religious labelled; Si adapter | Halves Whisper avg. WER on FLEURS-54 | FLEURS benchmark | Best published FLEURS-Sinhala WER; religious-domain bias limits conversational applicability |
| Bartelds et al. [54] (low-resource langs) | 2023 | wav2vec 2.0 + TTS augmentation | < 10h per language | 25.5% rel. WER reduction (Gronings) | Multiple minority langs | TTS augmentation most effective for very low-resource; transferable to Sinhala |
| Hou et al. [51] (MetaAdapter) | 2022 | Adapter-based PEFT on XLS-R | CommonVoice low-resource langs | Competitive to full FT | Multiple low-resource langs | 20× parameter reduction with matched WER; PEFT critical for Sinhala fine-tuning scalability |

** WER values for Whisper fine-tuning are community-reported on Hugging Face Hub and have not been independently peer-reviewed. XLSR-53 and IndicWav2Vec WERs are reported on test splits of OpenSLR-52 or FLEURS-Sinhala. All HMM-based WERs are on vocabulary-constrained test sets and are not directly comparable to large-vocabulary results.*

A caveat is necessary in reading the WER reductions in Table I as evidence for a single cause. Only one comparison in this table is controlled in the sense of holding architecture and corpus fixed while varying annotation quality alone: Gamage et al.’s 28.55%→23.38% improvement [27],[28], both using the same E2E LF-MMI architecture on the same underlying OpenSLR-52 recordings, with the only change being the corrected annotations from Warusawithana et al. [13] — an 18.1% relative reduction attributable specifically to data quality. The other large reductions in the table are confounded: the 23% DNN-HMM-vs-GMM-HMM gain [21] reflects an architecture change on a fixed corpus, and the 35% XLSR-53 gain [10] reflects a change in both architecture and pretraining data simultaneously. These are not commensurable with the Gamage figure, and none of them isolates corpus quality as a variable. We therefore restrict our claim that corpus quality is the dominant lever in Sinhala ASR to the one controlled comparison available in the literature, while noting that no study to date has held architecture fixed and varied corpus size or quality at multiple points — an omission closely related to Gap G2 (Section IX), and a natural next controlled experiment for the field.

## VIII. KEY CHALLENGES IN SINHALA ASR

### *A. Data Scarcity and Quality*

Data availability and quality are the most crucial obstacles to developing high-quality speech recognition capabilities in Sinhala. With only about 224 hours of speech data in the largest publicly available corpus (OpenSLR-52) [11], Sinhala lies some two orders of magnitude below the sizes of datasets commonly used for ASR of English or Mandarin. Crucially, this is entirely read data collected from one narrow dialect spoken by educated, urban Sinhalese people reading syntactically formal, phonetically well-balanced sentences. Regional variations in Central Province (Kandy dialect), Southern Province (Matara/Galle dialect), and Northern Sinhala go completely ignored.

The scarcity of training material is compounded by quality issues, as inconsistency in annotations in OpenSLR-52 was reported by Warusawithana et al. [13], resulting in a 15.9% relative WER penalty when considering it. The presence of annotation mistakes creates extra labeling noise, which negatively impacts the model's ability to distinguish fine-grained phonological features. Currently, no common preprocessing pipeline for Sinhala speech has been developed, and the lack of train/dev/test splits makes it impossible to conduct meaningful comparisons between different articles.

### *B. Morphological Complexity and High OOV Rates*

The presence of agglutinative morphology in Sinhala necessitates an open vocabulary; despite having 200 hours of training data, it's reasonable to expect some proportion of the forms in the test dataset not to be included in the training dataset, which would cause language models to assign them a probability of zero. In this case, the decoder will have to rely entirely on acoustic data when the likelihood of confusing the sounds reaches its maximum. Although this can be partially alleviated through subword tokenization, very complex morphological forms might still require unusual subword combinations [61].

Sinhala being in SOV order makes the problem with language models even more acute: the verb form, which usually contains a lot of information about the word, including markers for tense, aspect, and evidentiality, goes last, depriving the left-to-right language model of the syntactic context for the most complex parts of the sentence. This issue can be addressed using bidirectional language models, such as those in the BERT family [59], but integrating them with autoregressive models in ASR tasks is not straightforward and often requires rescoring approaches or non-autoregressive decoders.

### *C. Dialectal and Acoustic Variability*

Finally, Sinhala is an extremely variable language, ranging from the formal literary language to urban colloquial forms to rural dialects, as well as English-Sinhala code-switching; however, all of these variants remain unrecognized in ASR systems trained on read speech data. Phonological differences between the formal and colloquial variants of Sinhala are not merely stylistic: colloquial Sinhala undergoes systematic vowel reduction, consonant deletion, and morphological simplification that can change both the surface phoneme sequence and the word boundary placement of common utterances [3].

The lack of work on Sinhala code-switching ASR is particularly problematic, as mixed-language utterances are commonplace in Sri Lanka, where words from English and Sinhala are often used interchangeably, sometimes even within a single sentence. Inconsistencies in XLS-R/Whisper fine-tuning pipelines for multilingual ASR are based on the assumption that inputs should be monolingual, hence creating systematic errors in Sinhala with code-switched words.

### *D. Lack of Standardised Evaluation Benchmarks*

The lack of an openly shared Sinhala ASR benchmark with detailed data splits, evaluation protocol, and submission leaderboard makes comparison of various system implementations by competing research groups impossible. The WER range reported for Sinhala ASR is from 4% (for a small vocabulary, constrained task) to 45% (for a large vocabulary, broadcast ASR task) (Table I); however, no comparisons between these numbers can be made as they differ in terms of test sets used, scoring method, and data preprocessing procedure.

It would help a lot in making progress in Sinhala ASR to establish a benchmark dataset that is equivalent to LibriSpeech (clean read speech), TED-LIUM (prepared speech), or Switchboard (conversational speech). The FLEURS-Sinhala [12] dataset could serve as a basis for establishing a benchmark, as it contains a standardized 12-hour speech corpus with clear train/dev/test splits. Still, the nature of English-to-Sinhala translation datasets makes them less suitable for evaluating natural Sinhala speech.

### *E. Computational Resource Constraints*

Training large-scale pre-trained models, such as XLS-R-300M, requires multiple GPUs, which are not available at Sinhala-focused universities. There is a discrepancy in the Sinhala community regarding the availability of computational resources: the most efficient approaches require resources that are unavailable to the very researchers with extensive experience working with Sinhala language data and knowledge. While parameter-efficient fine-tuning approaches (LoRA, adapters) [51], [52] solve part of the problem, as they reduce fine-tuning computation requirements to a single GPU, the first stage of pre-training is still resource-intensive, and the pre-training is typically done using global pretrained models, which may be poorly adapted for Sinhala.

### *F. Real-Time Processing and Deployment Constraints*

ASR applications for live subtitles for Sri Lanka TV, interactive voice response systems (IVR), and hospital speech transcription require inference with low (<500 ms) latency on embedded hardware/cloud devices. Large transformer-based models with the best WER results have 1-3-second latency when run on CPU and high GPU memory requirements for batch inference. There has been little effort to apply model compression techniques such as knowledge distillation [65], quantization, and weight pruning, which have not been systematically explored for Sinhala ASR, and the trade-off between latency and WER for Sinhala models remains unknown.

## IX. RESEARCH GAPS

In light of the above analysis, the following represent some specific and actionable research gaps identified in the Sinhala ASR research literature:

G1: Lack of publicly available, annotated code-switched Sinhala–English speech corpus. Perhaps the most prominent gap in Sinhala NLP datasets relates to the lack of a publicly available, annotated Sinhala-English code-switched speech corpus. Publication of a corpus consisting of 20–30 hours of code-switched speech annotated on a per-utterance basis would enable the diagnosis of current models and the design of better ones.

G2: Lack of a standardized benchmark for large-vocabulary conversational Sinhala ASR. Whereas FLEURS-Sinhala provides a benchmark for Sinhala read speech ASR systems, there is a need for a benchmark involving conversational (domain-dependent and independent) dialectal speech. Four domains (healthcare, education, media, and governmental) and two dialects should be included in such a benchmark.

G3: Sinhala ASR Morphological tokenization is underresearched. While GPE [63] has been explored in a Sinhala machine translation context, this approach has not been explored for Sinhala ASR. A controlled experiment comparing the behavior of character, BPE, SentencePiece, and GPE tokenizers on Sinhala CTC fine-tuning would inform future architectures.

G4: PEFT solutions to Sinhala ASR have not been tested. Neither the adapter nor LoRA fine-tuning methods have been evaluated on Sinhala using peer-reviewed benchmarks (XLS-R and Whisper models). This makes it inaccessible to many researchers to run experiments in resource-constrained settings because of the computational complexity of fine-tuning.

G5: Hypothesis rescoring with LLM for Sinhala ASR not attempted. There are Sinhala-capable language models, such as SinBERT [66] and SinLlama [60], that can be employed for the Sinhala N-best rescoring task. However, no paper has reported combining a Sinhala ASR system with a Sinhala LLM to reduce WER at a marginal training cost.

G6: The noise resistance of Sinhala ASR systems has not been defined. None of the peer-reviewed articles reports the performance of any Sinhala model on noisy data. Particularly, no WER scores of MUSAN, WHAM!, or other corpora are available. In the absence of such benchmarks, one cannot forecast how these systems will perform in a real-world environment.

## X. FUTURE DIRECTIONS

### *A. Comprehensive Sinhala Speech Dataset Development*

Addressing the data challenge involves bringing together the community in the creation of five datasets: (1) a conversational Sinhala dataset with large vocabulary (target: 500 hours, 4 regions, 3 age groups); (2) a Sinhala-English code-switched dataset (target: 30 hours with language boundary annotations); (3) domain-specific corpora in healthcare, legal and education (target: 50 hours per domain); (4) noise augmented versions of current corpora at noise-to-signal ratios of 0, 5, 10, and 20 dB; (5) child-speech Sinhala dataset for educational application development.

Cost-effective crowdsourced collection facilities include: crowd-sourcing services such as Mozilla Common Voice, Appen, and the Voicer platform [42]. Linguistic community participation through Sri Lankan universities, broadcasting institutions, and government agencies will be crucial for gathering spontaneously produced data rather than being prompted. Quality control measures should follow the systematic protocol of Warusawithana et al. [13], with automatic quality screening followed by expert phonetician review of a random sample.

### *B. Language-Family SSL Pre-Training*

The better performance of IndicWav2Vec [48] on Sinhala than that of globally pre-trained XLS-R shows that pre-training within a language family is more efficient in data terms than global pre-training for low-resource languages. It follows that a South Asian Speech Foundation Model should be built by training a model on the concatenation of IndicWav2Vec's 17,000-hour South Asian corpus, OpenSLR-52, Mozilla Common Voice Sinhala, and all additional datasets collected according to direction A. Such a foundation model could reside on the Hugging Face Hub for further use in South Asian language technology.

### *C. Context-Aware Decoding with Sinhala LLMs*

The advancement of Sinhala language capable language models (SinBERT [66], SinLlama [60]) enables an N-best rescoring approach that can now be implemented for ASR but has not yet been considered. This approach: (1) acoustic wav2vec 2.0 / XLS-R model produces a list of hypotheses with a size of 50 with beam search; (2) scoring each hypothesis for grammatical soundness, morphological well-formedness, and domain-conditional perplexity using SinBERT or SinLlama; (3) minimum Bayes risk (MBR) combination of the final hypothesis. Such an approach does not entail joint training of language and acoustic models and can be composed using off-the-shelf resources, making it a high-priority near-term experiment.

### *D. Parameter-Efficient Adaptation and On-Device Deployment*

Low-rank adaptation (LoRA) fine-tuning of Whisper-small or XLS-R-300M on Sinhala should be systematically evaluated across the ranks of (r=4, 8, 16, 64) and modules to which LoRA is applied (Q/K/V attention layers, feed-forward layers, all layers) while minimizing the number of parameters subject to tuning and maximizing the WER score. LoRA-tuned PEFT models should be quantized to INT8 and INT4 using GPTQ or llm.int8() and benchmarked with respect to real-time factor (RTF) on ARM-based edge computing devices.

### *E. Standardised Benchmarking Infrastructure*

An open Sinhala ASR leaderboard system on par with Open ASR Leaderboard on Hugging Face [67] is required to address the following points: (1) provide a holdout secret test set that would reflect Sinhala speech under various conditions; (2) guarantee equal chances for comparison between methods by providing code for WER computation (case normalization, punctuation removal, and specific Sinhala characters normalization); (3) provide tests on clean and noise sets; (4) accept model parameters and code necessary for reproduction of the results.

## XI. CONCLUSION

Automatic Speech Recognition for Sinhala has a history that ranges from HMM-based systems in the early 2000s to current large, pre-trained, multilingual speech foundation models. Encouraging progress has been made: self-supervised pre-training, such as the XLS-R, IndicWav2Vec, Whisper, and MMS families, has brought Sinhala WER to below 30% (can be achieved with minimal fine-tuning on freely available data) and down to 15-25% (reachable with very limited fine-tuning on freely available data), though this range is community-reported and has not been independently peer-reviewed (see Table I note).

However, these improvements are still insufficient to fully solve the problem. First, the existing Sinhala speech dataset is small (insufficiently heterogeneous) and is ecologically invalid. Second, the published WER values cannot be compared across papers because they lacked standardized benchmarks. Third, the peculiarities of Sinhala morphology, i.e., the suffixation-based structure of the language and SOV syntax, are typically overlooked when the acoustically oriented SSL paradigm is prioritized in language modeling. Fourth, the difference between the laboratory WER and actual-world performance in the presence of background noise, dialectical variation, and domain shift is not quantified.

Overcoming these challenges requires concerted efforts across three areas: data, methods, and infrastructure. Regarding that, it requires systematic corpus construction aimed at conversational, dialectal, and noisy speech. As a method, it involves exploiting family-level pre-training of speech foundation models, parameter-efficient fine-tuning, family-embedding rescoring of speech foundation models, and morphological tokenization. Lastly, in the case of infrastructure, it needs to be standardized in benchmarking and open publication of models. The research program to be used in Sections IX and X is based on the specifics of the Sinhala language and on the effectiveness of a similarly motivated approach in a similar language.

Sinhala Automatic Speech Recognition is not a narrow technical question. It represents the broader issue of bridging speech technology with the planet's linguistically diverse populations, which increasingly rely on voice interfaces to access education, health care, public services, and economic opportunity. The methodological lessons learned in addressing the rich morphology, data scarcity, and dialectal diversity of Sinhala readily apply to dozens of other low-resource languages across South Asia, Southeast Asia, and Africa. We hope that our survey will be useful both as a comprehensive review of the Sinhala ASR literature and as a methodological manual for low-resource ASR researchers worldwide.

## REFERENCES

[1] R. Prabhavalkar, T. Hori, T.N. Sainath, R. Schlüter, and S. Watanabe, "End-to-end speech recognition: A survey," IEEE/ACM Trans. Audio, Speech, Lang. Process., vol. 32, pp. 325–351, 2024.

[2] S. Ranathunga, E.-S.A. Lee, M. Prifti Skenduli, R. Shekhar, M. Alam, and R. Kaur, "Neural machine translation for low-resource languages: A survey," ACM Comput. Surv., vol. 55, no. 11, pp. 1–37, 2023.

[3] N. de Silva, "Survey on publicly available Sinhala natural language processing tools and research," arXiv:1906.02358, 2019.

[4] L. A. D. S. A. Molligoda and P. G. Wijayarathna, "Applicability of Hidden Markov Model approach for Sinhala speech recognition: A systematic review," unpublished manuscript.

[5] C. Liyanage, R. Pushpananda, D.L. Herath, and R. Weerasinghe, "A computational grammar of Sinhala," in Proc. Int. Conf. Comput. Linguistics Intell. Text Process., 2012, pp. 188–200.

[6] R. Weerasinghe, D. Herath, and V. Welgama, "Corpus-based Sinhala lexicon," in Proc. 7th Workshop Asian Lang. Resources (ALR7), 2009, pp. 17–23.

[7] T. Nadungodage, C. Liyanage, A. Perera, R. Pushpananda, and R. Weerasinghe, "Sinhala G2P conversion for speech processing," in Proc. 6th Workshop Spoken Lang. Technol. Under-Resourced Lang. (SLTU), 2018, pp. 112–116.

[8] A. Wasala, R. Weerasinghe, and K. Gamage, "Sinhala grapheme-to-phoneme conversion and rules for schwa epenthesis," in Proc. COLING/ACL 2006 Poster Sessions, 2006, pp. 890–897.

[9] T. Dinushika, L. Kavmini, P. Abeyawardhana, U. Thayasivam, and S. Jayasena, "Speech command classification system for Sinhala language based on automatic speech recognition," in Proc. Int. Conf. Asian Lang. Process. (IALP), 2019, pp. 205–210.

[10] A. L. Nanayakkara and R. Weerasinghe, "Exploring model-level transfer learning for improving Sinhala speech recognition," DOI: 10.13140/RG.2.2.28491.77602.

[11] O. Kjartansson, S. Sarin, K. Pipatsrisawat, M. Jansche, and L. Ha, "Crowd-sourced speech corpora for Javanese, Sundanese, Sinhala, Nepali, and Bangladeshi Bengali," in Proc. 6th SLTU Workshop, 2018, pp. 52–55.

[12] A. Conneau, M. Ma, S. Khanuja, Y. Zhang, V. Axelrod, S. Dalmia, J. Riesa, C. Rivera, and A. Bapna, "FLEURS: Few-shot learning evaluation of universal representations of speech," in Proc. IEEE SLT, 2022.

[13] D. Warusawithana, N. Kulaweera, L. Weerasinghe, and B. Karunarathne, "A systematic approach to derive a refined speech corpus for Sinhala," in Proc. LREC 2022, pp. 5107–5113.

[14] L. Rabiner and B.-H. Juang, "An introduction to hidden Markov models," IEEE ASSP Mag., vol. 3, no. 1, pp. 4–16, 1986.

[15] S. Young et al., The HTK Book, Cambridge University Engineering Department, 2006.

[16] D. Huggins-Daines, M. Kumar, A. Chan, A. W. Black, M. Ravishankar, and A. I. Rudnicky, "PocketSphinx: A free, real-time continuous speech recognition system for hand-held

devices," in Proc. IEEE ICASSP, 2006, pp. I-185–I-188, DOI: 10.1109/ICASSP.2006.1659988.

[17] D. Povey et al., "The Kaldi speech recognition toolkit," in Proc. IEEE ASRU Workshop, 2011.

[18] G.E. Hinton et al., "Deep neural networks for acoustic modeling in speech recognition," IEEE Signal Process. Mag., vol. 29, no. 6, pp. 82–97, 2012.

[19] V. Peddinti, D. Povey, and S. Khudanpur, "A time delay neural network architecture for efficient modeling of long temporal contexts," in Proc. Interspeech, 2015, pp. 3214–3218.

[20] A. Graves, A.-R. Mohamed, and G. Hinton, "Deep recurrent neural networks for acoustic modelling," in Proc. IEEE ICASSP, 2013, pp. 6645–6649.

[21] N.A.K.H.S. Karunathilaka, V. Welgama, T. Nadungodage, and R. Weerasinghe, "Low-resource Sinhala speech recognition using deep learning," in Proc. IEEE ICTer, 2020, pp. 196–201.

[22] A. Graves, S. Fernández, F. Gomez, and J. Schmidhuber, "Connectionist temporal classification: Labelling unsegmented sequence data with recurrent neural networks," in Proc. ICML, 2006, pp. 369–376.

[23] W. Chan, N. Jaitly, Q.V. Le, and O. Vinyals, "Listen, attend and spell," in Proc. IEEE ICASSP, 2016.

[24] A. Graves, "Sequence transduction with recurrent neural networks," arXiv:1211.3711, 2012.

[25] L. Dong, S. Xu, and B. Xu, "Speech-Transformer: A no-recurrence sequence-to-sequence model for speech recognition," in Proc. IEEE ICASSP, 2018.

[26] A. Gulati et al., "Conformer: Convolution-augmented Transformer for speech recognition," in Proc. Interspeech, 2020, pp. 5036–5040.

[27] B. Gamage, R. Pushpananda, T. Nadungodage, and R. Weerasinghe, "Improve Sinhala speech recognition through e2e LF-MMI model," in Proc. ICON 2021, ACL Anthology, 2021.

[28] B. Gamage, R. Pushpananda, T. Nadungodage, and R. Weerasinghe, "Applicability of end-to-end deep neural architecture to Sinhala speech recognition," Int. J. Adv. ICT Emerg. Reg. (ICTer), vol. 17, no. 1, pp. 17–21, 2024.

[29] S. Schneider, A. Baevski, R. Collobert, and M. Auli, "wav2vec: Unsupervised pre-training for speech recognition," in Proc. Interspeech, 2019, pp. 3465–3469.

[30] A. Baevski, S. Schneider, and M. Auli, "vq-wav2vec: Self-supervised learning of discrete speech representations," in Proc. ICLR, 2020.

[31] A. Baevski, H. Zhou, A. Mohamed, and M. Auli, "wav2vec 2.0: A framework for self-supervised learning of speech representations," in Proc. NeurIPS, 2020.

[32] W.-N. Hsu, B. Bolte, Y.-H.H. Tsai, K. Lakhotia, R. Salakhutdinov, and A. Mohamed, "HuBERT: Self-supervised speech representation learning by masked prediction of hidden units," IEEE/ACM Trans. Audio, Speech, Lang. Process., vol. 29, pp. 3451–3460, 2021.

[33] S. Chen et al., "WavLM: Large-scale self-supervised pre-training for full stack speech processing," IEEE J. Sel. Top. Signal Process., vol. 16, no. 6, pp. 1505–1518, 2022.

[34] A. Babu et al., "XLS-R: Self-supervised cross-lingual speech representation learning at scale," in Proc. Interspeech, 2022. arXiv:2111.09296.

[35] A. Conneau, A. Baevski, R. Collobert, A. Mohamed, and M. Auli, "Unsupervised cross-lingual representation learning for speech recognition," in Proc. Interspeech, 2021.

[36] A. Radford, J.W. Kim, T. Xu, G. Brockman, C. McLeavey, and I. Sutskever, "Robust speech recognition via large-scale weak supervision," in Proc. ICML, PMLR 202, pp. 28492–28518, 2023.

[37] V. Pratap et al., "Scaling speech technology to 1,000+ languages," J. Mach. Learn. Res. (JMLR), vol. 25, 2024. arXiv:2305.13516.

[38] T. Nadungodage, V. Welgama, and R. Weerasinghe, "Developing a speech corpus for Sinhala speech recognition," in Proc. 10th Int. Conf. Natural Language Processing (ICON-2013), Noida, India, Dec. 2013.

[39] T. Nadungodage and R. Weerasinghe, "Continuous Sinhala speech recognizer," in Proc. HLTD, 2011, pp. 141–147.

[40] T. Nadungodage, R. Weerasinghe, and M. Niranjan, "Speaker adaptation applied to Sinhala speech recognition," Int. J. Comput. Linguist. Appl. (IJCLA), 2013.

[41] K. Sodimana, P. De Silva, S. Sarin, O. Kjartansson, M. Jansche, K. Pipatsrisawat, and L. Ha, "A step-by-step process for building TTS voices for Bangla, Javanese, Khmer, Nepali, Sinhala, and Sundanese," in Proc. 6th SLTU Workshop, 2018, pp. 66–70.

[42] D. Buddhika, R. Liyadipita, S. Nadeeshan, H. Witharana, S. Jayasena, and U. Thayasivam, "Domain specific intent classification of Sinhala speech data," in Proc. IEEE IALP, 2018, pp. 197–202.

[43] Y. Karunanayake, U. Thayasivam, and S. Ranathunga, "Transfer learning based free-form speech command classification for low-resource languages," in Proc. ACL 2019 SRW, pp. 288–294.

[44] T. Dinushika, L. Kavmini, P. Abeyawardhana, U. Thayasivam, and S. Jayasena, "Improved speech command classification system for Sinhala language based on automatic speech recognition," Int. J. Asian Lang. Process. (IJALP), vol. 30, 2020.

[45] W. Manamperi, D. Karunathilake, T. Madhushani, N. Galagedara, and D. Dias, "Sinhala speech recognition for interactive voice response systems accessed through mobile phones," in Proc. IEEE MERCon, 2018, pp. 241–246.

[46] R.V.P.S. Akesh and R.G.N. Meegama, "Real-time subtitle generator for Sinhala speech," unpublished.

[47] S. Watanabe et al., "ESPnet: End-to-end speech processing toolkit," in Proc. Interspeech, 2018. arXiv:1804.00015.

[48] T. Javed et al., "Towards building ASR systems for the next billion users (IndicWav2Vec)," in Proc. AAAI, 2022. arXiv:2111.03945.

[49] D.S. Park, W. Chan, Y. Zhang, C.-C. Chiu, B. Zoph, E.D. Cubuk, and Q.V. Le, "SpecAugment: A simple data augmentation method for automatic speech recognition," in Proc. Interspeech, 2019, pp. 2613–2617.

[50] J. Valk and T. Alumäe, "VoxLingua107: A dataset for spoken language recognition," in Proc. IEEE SLT, 2021. arXiv:2011.12998.

[51] W. Hou, H. Zhu, Y. Wang, J. Wang, T. Qin, R. Xu, and T. Shinozaki, "Exploiting adapters for cross-lingual low-resource speech recognition," IEEE/ACM Trans. Audio, Speech, Lang. Process., vol. 30, pp. 317–329, 2022.

[52] E.J. Hu et al., "LoRA: Low-rank adaptation of large language models," in Proc. ICLR, 2022. arXiv:2106.09685.

[53] J. Mainzinger and G.-A. Levow, "Fine-tuning ASR models for very low-resource languages: A study on Mvskoke," in Proc. ACL SRW, 2024.

[54] M. Bartelds, N. San, B. McDonnell, D. Jurafsky, and M. Wieling, "Improving low-resource ASR by TTS-based augmentation and self-training," in Proc. ACL, 2023.

[55] D.S. Park, Y. Zhang, Y. Jia, W. Han, C.-C. Chiu, B. Li, Y. Wu, and Q.V. Le, "Improved noisy student training for automatic speech recognition," in Proc. Interspeech, 2020.

[56] Y. Zhang et al., "Google USM: Scaling automatic speech recognition beyond 100 languages," arXiv:2303.01037, 2023.

[57] A. Sriram, H. Jun, S. Satheesh, and A. Coates, "Cold fusion: Training seq2seq models together with language models," in Proc. Interspeech, 2018.

[58] R. Cabrera, X. Liu, M. Ghodsi, Z. Matteson, E. Weinstein, and A. Kannan, "Language model fusion for streaming end-to-end speech recognition," arXiv:2104.04487, 2021.

[59] C. Chen, Y. Hu, C.-H.H. Yang, S.M. Siniscalchi, P.-Y. Chen, and E.-S. Chng, "HyPoradise: An open baseline for generative speech recognition with large language models," in Proc. NeurIPS D&B, 2023.

[60] H.W.K. Aravinda, R. Sirajudeen, S. Karunathilake, N. de Silva, R. Kaur, A.S. Bhankhar, and S. Ranathunga, "SinLlama: A large language model for Sinhala," arXiv:2508.09115, 2025.

[61] T. Kudo and J. Richardson, "SentencePiece: A simple and language-independent subword tokenizer and detokenizer for neural text processing," in Proc. EMNLP Demos, 2018.

[62] K. Manohar, A. R. Jayan, and R. Rajan, "Improving speech recognition systems for the morphologically complex Malayalam language using subword tokens for language modeling," EURASIP J. Audio Speech Music Process., vol. 2023, art. no. 47, 2023, DOI: 10.1186/s13636-023-00313-7.

[63] Velayuthan et al., "Grapheme pair encoding (GPE): A subword tokenization strategy for complex scripts," in Proc. Interspeech, 2024.

[64] G. Pundak, T.N. Sainath, R. Prabhavalkar, A. Kannan, and D. Zhao, "Deep context: End-to-end contextual speech recognition," in Proc. IEEE SLT, 2018.

[65] G. Hinton, O. Vinyals, and J. Dean, "Distilling the knowledge in a neural network," arXiv:1503.02531, 2015.

[66] V. Dhananjaya, P. Demotte, S. Ranathunga, and S. Jayasena, "BERTifying Sinhala: A comprehensive analysis of pre-trained language models for Sinhala text classification," in Proc. LREC, 2022, pp. 7377–7385.

[67] Open ASR Leaderboard, Hugging Face Hub, 2024. [Online]. Available: https://huggingface.co/spaces/hf-audio/open_asr_leaderboard

[68] P. G. N. Priyadarshani, N. G. J. Dias, and A. Punchihewa, "Dynamic time warping based speech recognition for isolated Sinhala words," in Proc. IEEE 55th Int. Midwest Symp. Circuits Syst. (MWSCAS), 2012, pp. 892–895, DOI: 10.1109/MWSCAS.2012.6292164.